\documentclass[runningheads]{llncs}
\usepackage[T1]{fontenc}
\usepackage{graphicx,verbatim}
\usepackage{booktabs}
\usepackage{amsmath, amssymb, graphicx} 
\usepackage{multirow}
\begin{document}

\title{Contributing Title}
%
\titlerunning{AdFair-CLIP: Fair Contrastive Learning for CXR Diagnosis}
\title{AdFair-CLIP: Adversarial Fair Contrastive Language-Image Pre-training for Chest X-rays}
\author{Chenlang Yi\inst{1} \and Zizhan Xiong\inst{1} \and Qi Qi\inst{2} \and Xiyuan Wei\inst{1} \and Girish Bathla\inst{3} \and Ching-Long Lin\inst{2} \and Bobak J. Mortazavi\inst{1} \and Tianbao Yang\inst{1}\thanks{Corresponding author.}}

%
\authorrunning{C. Yi et al.}
\institute{
Texas A\&M University, College Station, TX, USA \\
\email{\{baileyyeah, tianbao-yang\}@tamu.edu} \and
The University of Iowa, Iowa City, IA, USA \and
Mayo Clinic School of Medicine, Rochester, MN, USA
}

\maketitle              
\begin{abstract}
Contrastive Language-Image Pre-training (CLIP) models have demonstrated superior performance across various visual tasks including medical image classification. However, fairness concerns, including demographic biases, have received limited attention for CLIP models. This oversight leads to critical issues, particularly those related to race and gender, resulting in disparities in diagnostic outcomes and reduced reliability for underrepresented groups. To address these challenges, we introduce AdFair-CLIP, a novel framework employing adversarial feature intervention to suppress sensitive attributes, thereby mitigating spurious correlations and improving prediction fairness. We conduct comprehensive experiments on chest X-ray (CXR) datasets, and show that AdFair-CLIP significantly enhances both fairness and diagnostic accuracy, while maintaining robust generalization in zero-shot and few-shot scenarios. These results establish new benchmarks for fairness-aware learning in CLIP-based medical diagnostic models, particularly for CXR analysis.
\end{abstract}
\section{Introduction}  
As artificial intelligence (AI) systems increasingly inform clinical decision-making, concerns have arisen over biases in AI algorithms~\cite{seyyed2021underdiagnosis,obermeyer2019dissecting,jin2024fairmedfm}. These biases can disproportionately disadvantage underrepresented populations, reinforcing healthcare disparities. Ensuring fairness in medical AI is crucial to mitigating demographic biases and promoting equitable healthcare outcomes.

CLIP models, integrating visual and textual modalities, show promise in medical AI for disease diagnosis and report generation~\cite{huang2023visual,zhang2022contrastive,ghosh2024clipsyntel,zhang2025biomedclipmultimodalbiomedicalfoundation,qi2020simpleeffectiveframeworkpairwise}. In the CXR domain, models like GLoRIA~\cite{huang2021gloria}, CXR-CLIP~\cite{you2023cxr}, and MedCLIP~\cite{wang2022medclip} have achieved strong diagnostic performance via contrastive learning on large image-text datasets. However, prior fairness research has focused on vision-only models~\cite{glocker2023algorithmic,khan2023fair,seyyed2020chexclusion,brown2022detecting,qi2024provableoptimizationadversarialfair}, while fairness in CLIP-based methods remains largely unexplored. Ensuring fairness in these models is essential to preventing biased outcomes, underscoring the need for a comprehensive fairness investigation. Existing fairness-aware methods often face trade-offs or limited generalizability. CLIP-clip~\cite{wang2021genderneutralqueriesreallygenderneutral} mitigates bias by pruning bias-associated dimensions but inadvertently removes task-relevant features. FairCLIP~\cite{luo2024fairclip} introduces similarity-level interventions via Sinkhorn divergence minimization. However, these adjustments remain superficial and fail to fully eliminate biases in feature representations. DebiasCLIP~\cite{berg2022promptarraykeepsbias} employs a frozen backbone strategy, modifying only cross-modal similarity scores. While effective for zero-shot classification, this approach lacks adaptability to downstream tasks, limiting its debiasing impact.

To overcome these limitations, we propose AdFair-CLIP, an adversarial fairness framework that integrates CLIP~\cite{radford2021learning} with adversarial learning to mitigate demographic biases while preserving multimodal representation learning. Unlike prior methods, AdFair-CLIP operates directly in the feature space, formulates bias mitigation as a minimax optimization problem~\cite{ganin2016domain}, and reduces spurious correlations between demographic attributes and model predictions, allowing the model to focus on disease-relevant features. We evaluate our approach on CheXpert Plus~\cite{chambon2024chexpertplusaugmentinglarge} and MIMIC-CXR~\cite{johnson2019mimic}, assessing fairness across race and gender. Experimental results demonstrate substantial improvements in both fairness and diagnostic accuracy under zero-shot and few-shot conditions.

Our main contributions are as follows: (1) We conduct a systematic fairness investigation of state-of-the-art (SOTA) CLIP models for CXR diagnosis, revealing significant biases across demographics. (2) We propose AdFair-CLIP, a novel framework employing adversarial feature-space intervention to mitigate these biases and address fairness in CLIP-based CXR diagnostic models. (3) We demonstrate AdFair-CLIP's effectiveness through comprehensive evaluations, showing substantial improvements in both fairness and diagnostic accuracy.

\section{Method} 

\subsection{Fairness Investigation}
Dataset imbalances in CXR analysis can introduce biases~\cite{brown2022detecting,zhang2022improving}, leading CLIP-based models to develop biased representations. We take racial bias as an example, which stems from two key sources. First, CXR datasets predominantly consist of White patients, with Black and Asian populations underrepresented. As a result, models learn more informative features for the majority group while extracting less meaningful representations for underrepresented groups, resulting in predictive disparities. Second, disease prevalence differs across racial groups. In the CheXpert Plus~\cite{chambon2024chexpertplusaugmentinglarge} dataset, Cardiomegaly is more prevalent in Asian patients than in Black and White patients. This imbalance may cause models to associate disease likelihood with racial features rather than pathology, reinforcing biases. These issues affect both zero-shot inference and downstream tasks, producing diagnostic disparities. To systematically evaluate these concerns, we analyze SOTA CLIP-based CXR diagnostic models using multiple fairness metrics, uncovering significant biases (detailed in Experiment \& Analysis).

\subsection{AdFair-CLIP}
\begin{figure}[tb]
    \centering
    \includegraphics[width=0.91\linewidth]{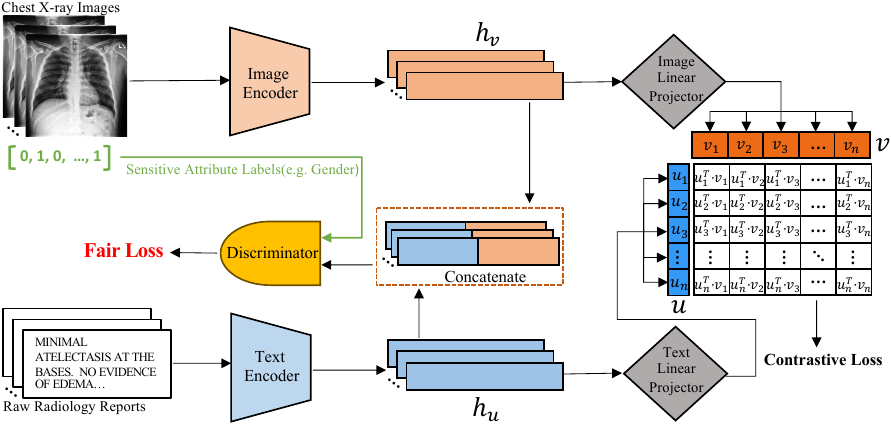}
    \caption{Overview of the AdFair-CLIP architecture. Image and text encoders extract representations $h_v$ and $h_u$ from chest X-ray images and radiology reports, projecting them to $v$ and $u$ for contrastive alignment. A discriminator predicts sensitive attributes from concatenated representations, enabling adversarial training to mitigate biases.}
    \label{fig:afclip-architecture}
\end{figure}

To address fairness challenges, we propose AdFair-CLIP, a min-max adversarial optimization framework that mitigates demographic bias while preserving multimodal alignment. The overview of the method is shown in Fig.~\ref{fig:afclip-architecture}.

{\bf Multi-Modal Alignment.}  
We adopt ResNet50~\cite{he2016deep} as the image encoder $E_v(\cdot)$ and BioClinicalBERT~\cite{alsentzer2019publicly} as the text encoder $E_t(\cdot)$. Given a chest X-ray image $I$ and a radiology report $T$, the encoders extract intermediate features $h_v = E_v(I) \in \mathbb{R}^{2048}$ and $h_u = \text{MeanPool}(E_t(T), M) \in \mathbb{R}^{768}$. These are projected into a shared space and $\ell_2$-normalized to obtain embeddings $v, u \in \mathbb{R}^{512}$. To align image and text representations, we employ the InfoNCE loss~\cite{oord2018representation}. The loss function $\mathcal{L}_{\text{GCL}}(w_v, w_u)$ is defined as follows:
\begin{equation}
-\frac{1}{N} \sum_{i=1}^{N} \left[ \log \frac{\exp\big(\mathrm{sim}(v_i, u_i)/\tau\big)}{\sum_{j=1}^{N} \exp\big(\mathrm{sim}(v_i, u_j)/\tau\big)} + \log \frac{\exp\big(\mathrm{sim}(u_i, v_i)/\tau\big)}{\sum_{j=1}^{N} \exp\big(\mathrm{sim}(u_i, v_j)/\tau\big)} \right],
\end{equation}
where \(w_v\) and \(w_u\) are the parameters of the vision and text encoders, respectively, \(\text{sim}(\cdot,\cdot)\) denotes cosine similarity, and $\tau$ is a learnable temperature parameter.

{\bf Fair Loss.}  
We design a discriminator to predict sensitive attributes. It is a multi-layer perceptron denoted as \(D(\cdot)\) that processes the concatenated intermediate representations \(h_v \in \mathbb{R}^{2048}\) and \(h_u \in \mathbb{R}^{768}\). The combined vector \(h_{\text{concat}} = [h_v; h_u] \in \mathbb{R}^{2816}\) is passed through \(D(\cdot)\), producing an output \(s = D(h_{\text{concat}}) \in \mathbb{R}^{C}\), where \(C\) is the number of sensitive attribute categories. To address sample imbalance in sensitive attributes, particularly for underrepresented groups, a weighted fair entropy loss is applied to ensure fair classification across all categories. The fairness loss is computed over a batch of $N$ samples, based on the discriminator's output logits $\mathbf{s}^{(k)} = [s_1^{(k)}, \dots, s_C^{(k)}] \in \mathbb{R}^{C}$ for each sample $k$. Let the ground-truth one-hot vector for the $k$-th sample be $\mathbf{y}^{(k)} = [y_1^{(k)}, \dots, y_C^{(k)}] \in \mathbb{R}^{C}$. The class weights $\mathbf{w} = [w_1, \dots, w_C]$, where $w_i = 1 / n_i$ and $n_i$ denotes the number of samples in category $i$ in a batch, are used to balance sample disparities. The weighted fair loss is defined as:
\begin{equation} \label{eq:fairness_loss}
\mathcal{L}_{\text{Fair}}(w_v, w_u, w_{\text{d}}) = - \frac{1}{N} \sum_{k=1}^{N} \sum_{i=1}^{C} w_i y_i^{(k)} \log \left( \frac{\exp(s_i^{(k)})}{\sum_{j=1}^{C} \exp(s_j^{(k)})} \right),
\end{equation}
where $w_\text{d}$ denotes the parameter of the discriminator network.

{\bf Adversarial Optimization.}
Our method is based on adversarial training, formulated as a minimax optimization problem \cite{ganin2016domain}. The discriminator aims to maximize the accuracy of predicting sensitive attributes from the learned representations, while the encoders seek to minimize the contrastive loss and reduce the accuracy of predicting sensitive attributes. This encourages the encoder to generate representations invariant to sensitive attributes. The final optimization objective is expressed as:
\begin{equation} \label{eq:final_objective}
    \min_{w_v, w_u} \max_{w_{\text{d}}} \mathcal{L}_{\text{GCL}}(w_v, w_u) - \alpha \cdot \mathcal{L}_{\text{Fair}}(w_v, w_u, w_{\text{d}}),
\end{equation}
where the hyperparameter $\alpha$ balances the trade-off between multimodal alignment and fairness regularization.

\section{Experiment \& Analysis}

\subsection{Datasets}
We use CheXpert Plus~\cite{chambon2024chexpertplusaugmentinglarge} and MIMIC-CXR~\cite{johnson2019mimic} for training, and evaluate exclusively on a demographically balanced test set sampled from CheXpert Plus. 

{\bf CheXpert Plus and MIMIC-CXR.}  CheXpert Plus extends CheXpert~\cite{irvin2019chexpert} for multimodal learning and fairness research, providing CXR images with radiology reports. We use the "impression" section as textual input, pre-train on frontal image-text pairs, and reserve expert-annotated samples for validation. We focus on frontal chest radiographs with 191,071 image-text pairs, using 3,000 randomly sampled training images for validation. MIMIC-CXR is a large-scale CXR dataset with radiology reports. We extract the "findings" and "impression" sections as text, pre-train on the training split, and retain a subset for validation. Further details are available in~\cite{zhang2022contrastive}. Both datasets include demographic metadata, enabling fairness assessment in model performance.

{\bf Fair Test Dataset.}
Previous CLIP-based models for CXR diagnosis have primarily used the CheXpert 5x200~\cite{huang2021gloria} dataset for evaluation, which underrepresents certain demographic groups. To ensure a fair evaluation, we construct a demographically balanced test set sampled from CheXpert Plus, explicitly stratified by race and gender. However, balancing demographic representation while adhering to the strict criteria in~\cite{huang2021gloria}, which require each image to contain positive labels for only one condition, limits the test dataset to 450 images across five disease categories (90 per class). This balanced test dataset enables a more rigorous assessment of models' fairness and is named FCXP 5x90.

\subsection{Baseline}
Our baselines span multiple model categories. To assess fairness in CLIP-based CXR diagnostic methods, we compare GLoRIA~\cite{huang2021gloria}, an attention-based framework for learning global and local representations, CXR-CLIP~\cite{you2023cxr}, mitigating data scarcity via synthetic image-text generation and multi-view contrastive supervision, and MedCLIP~\cite{wang2022medclip}, leveraging unpaired data and medical knowledge to reduce false negatives. Additionally, we include the original CLIP~\cite{radford2021learning} for comparison. To validate AdFair-CLIP's effectiveness, we benchmark it against two fairness-aware CLIP methods: FairCLIP~\cite{luo2024fairclip} and DebiasCLIP~\cite{berg2022promptarraykeepsbias}.

\subsection{Experimental Setup}
In this section, we outline the pre-training and evaluation strategies, as well as the metrics used to thoroughly assess predictive accuracy and fairness.

{\bf Pre-Training and Evaluation.} All models are independently pre-trained on CheXpert Plus and MIMIC-CXR, with image encoders initialized from ResNet-50~\cite{he2016deep} and text encoders from BioClinicalBERT~\cite{alsentzer2019publicly}, following the official checkpoints. To systematically evaluate model performance, we assess each pre-trained image encoder under two settings: zero-shot classification and few-shot classification. In the zero-shot setting, classification is framed as an image-text similarity task, where labels are converted into textual prompts following~\cite{huang2021gloria}. For few-shot classification, we freeze the pre-trained image encoder and train a linear classifier on top of it using only 10\% of the labeled training data from CheXpert Plus. All evaluations are conducted on FCXP 5x90, assessing fairness across race (Black, White, Asian) and gender (Male, Female). This setup allows us to assess both the models' performance on the in-domain and out-of-domain generalization.

{\bf Metrics.} To thoroughly evaluate the effectiveness of our proposed method, we employ both performance and fairness metrics. For performance evaluation, we use the accuracy and Area Under the Curve (AUC). To assess fairness, we incorporate multiple fairness-aware metrics. Demographic Parity Difference (DPD)~\cite{agarwal2018reductions} measures the maximum absolute difference in the probabilities of positive predictions across groups. Difference in Equalized Odds (DEOdds)~\cite{agarwal2018reductions} quantifies the maximum disparity in True Positive Rates (TPR) and False Positive Rates (FPR) separately across groups. Additionally, Group-wise AUC (GAUC)~\cite{yao2023stochastic} measures the probability that a randomly selected sample from one demographic group receives a higher predicted score than a randomly selected sample from another. Inter-Group AUC (Inter-AUC)~\cite{beutel2019fairnessrecommendationrankingpairwise} quantifies the degree to which the ranking of positive samples from one demographic group relative to negative samples from another group remains independent of group membership. Intra-Group AUC (Intra-AUC)~\cite{beutel2019fairnessrecommendationrankingpairwise} evaluates the consistency of ranking within each demographic group by comparing the ranking of positive and negative samples within the same group.

\textbf{Implementation Details.}  
Chest X-ray images are resized to \(256 \times 256\) and cropped to \(224 \times 224\). Radiology reports are tokenized using BioClinicalBERT~\cite{alsentzer2019publicly}. Pre-training uses the Adam optimizer~\cite{diederik2014adam} with a learning rate of \(1 \times 10^{-5}\), batch size 36, and weight decay \(1 \times 10^{-6}\), adjusted by ReduceLROnPlateau. The contrastive loss temperature \(\tau\) is 0.1, and the fairness regularization coefficient \(\alpha\) is 0.3. For linear evaluation, the learning rate is \(1 \times 10^{-4}\) with batch size 64. All experiments are implemented in PyTorch on six NVIDIA A100 GPUs.

\begin{figure}[tb]
    \centering
    \includegraphics[width=1\linewidth]{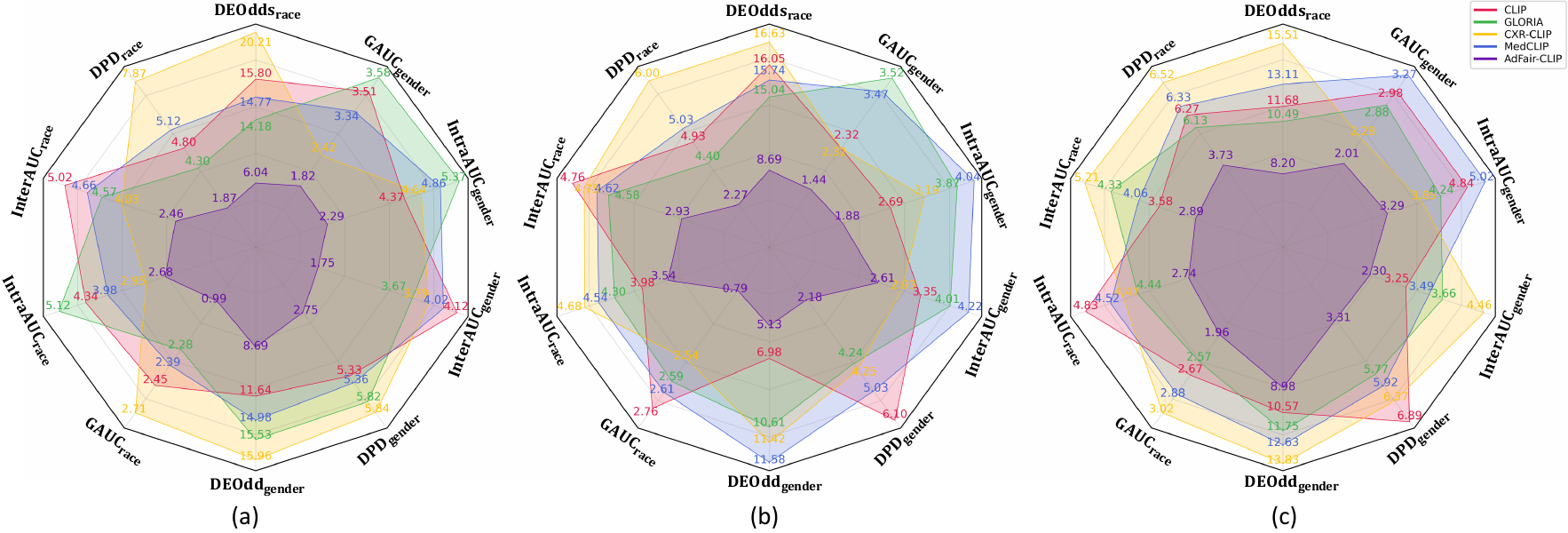}
    \caption{Fairness assessment of SOTA CLIP-based CXR diagnostic methods across race and gender in three scenarios, with all scores presented in percentage.}

    \label{fig:radar}
\end{figure}

\begin{figure}[tb]
    \centering
    \includegraphics[width=1\linewidth]{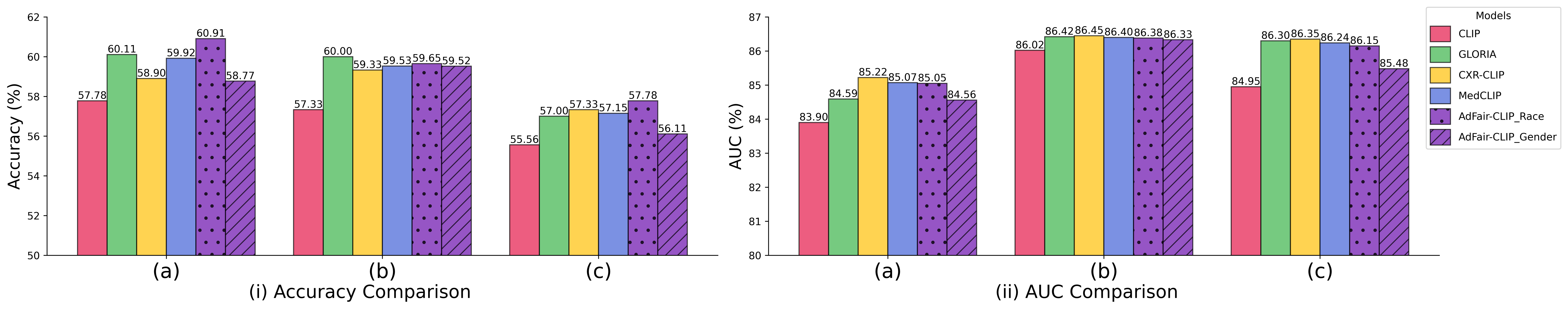}
    \caption{Performance assessment of SOTA CLIP-based CXR diagnostic methods across race and gender in three scenarios, with all scores presented in percentage.}
    \label{fig:bar}
\end{figure}

\subsection{Results Analysis}  
We evaluate three scenarios: (a) Zero-shot: models pre-trained on CheXpert Plus, (b) Few-shot: models pre-trained on CheXpert Plus and fine-tuned on 10\% of its data, and (c) Transfer learning: models pre-trained on MIMIC-CXR and fine-tuned on 10\% of the labeled data from CheXpert Plus. All results are averaged over three random seeds, with standard deviations ranging from 0.22\% to 5.18\%.

{\bf AdFair-CLIP vs. CLIP-based CXR Diagnostic Methods.} The fairness results in Fig.~\ref{fig:radar} reveal significant and persistent biases in CLIP-based CXR diagnostic models across race and gender groups. Each axis represents a fairness metric for race or gender, where lower values indicate better fairness and a larger enclosed area signifies greater disparities. GLoRIA, CXR-CLIP, MedCLIP, and CLIP exhibit substantial fairness gaps in multiple metrics, demonstrating limited effectiveness in ensuring equitable predictions. These biases persist across all evaluation scenarios, underscoring the need for stronger fairness interventions in medical AI. In contrast, AdFair-CLIP consistently achieves lower disparities across all metrics, effectively addressing fairness limitations in existing models. Moreover, Fig.~\ref{fig:bar} compares accuracy and AUC. AdFair-CLIP with racial bias mitigation performs competitively with SOTA baselines and even surpasses them in some cases, while its gender bias counterpart shows slightly lower performance yet remains superior to standard CLIP.
\begin{table}[t] 
    \centering
    \caption{Zero-shot comparison of AdFair-CLIP with baselines, pre-trained on CheXpert Plus, with all scores presented in percentage.}

    {\fontsize{8pt}{10pt}\selectfont
    
    \begin{tabular}{|c|l|c|c|c|c|c|c|c|}
        \hline
        \textbf{Attribute}& \textbf{Method} & \textbf{Acc} & \textbf{AUC} & \textbf{DPD} & \textbf{DEOdds} & \textbf{GAUC} & \textbf{Inter-AUC} & \textbf{Intra-AUC}\\\hline
        
        \multirow{4}{*}{\centering Race}
        & CLIP  & 57.78 & 83.90 & 4.80 & 15.80 & 2.45 & 5.02 & 4.34 \\
        & FairCLIP  & 60.89 & \textbf{85.08} & 3.20 & 8.99 & 1.64 & 3.40 & 3.15 \\
        & DebiasCLIP & 59.33 & 84.61 & 2.40 & 9.65 & 1.73 & 3.41 & 2.87 \\\cline{2-9}
        & AdFair-CLIP & \textbf{60.91} & 85.05 & \textbf{1.87} & \textbf{6.04}& \textbf{0.99} & \textbf{2.46} & \textbf{2.68}\\
        \hline
        \multirow{4}{*}{\centering Gender}
        & CLIP  & 57.78 & 83.90 & 5.33 & 11.64 & 3.51 & 4.12 & 4.37 \\
        & FairCLIP  & 58.34 & 84.37 & 3.12 & 10.28 & 2.51 & 3.29 & 3.27 \\
        & DebiasCLIP & 56.00 & 84.01 & 3.67 & 10.00 & \textbf{1.67}& 1.83 & 2.34 \\\cline{2-9}
        & AdFair-CLIP & \textbf{58.77} & \textbf{84.56} & \textbf{2.75} & \textbf{8.69} & 1.82 & \textbf{1.75} & \textbf{2.29}\\
        
        \hline
        
    \end{tabular}
    }
    
    \label{tab:chexpert_performance_metrics_zeroshot}
\end{table}
\begin{table}[tb]
    \centering
    \caption{Few-shot (10\%) comparison of AdFair-CLIP with baselines, pre-trained and fine-tuned on CheXpert Plus, with all scores presented in percentage.}

    {\fontsize{8pt}{10pt}\selectfont
    \begin{tabular}{|c|l|c|c|c|c|c|c|c|}
    \hline
    \textbf{Attribute} & \textbf{Method} & \textbf{Acc} & \textbf{AUC} & \textbf{DPD} & \textbf{DEOdds} & \textbf{GAUC} & \textbf{Inter-AUC} & \textbf{Intra-AUC}\\
    \hline
    \multirow{3}{*}{Race} 
        & CLIP  & 57.33 & 86.02 & 4.93 & 16.05 & 2.76 & 4.76 & 3.98 \\
        & FairCLIP  & 59.55 & 86.24 & 3.03 & 10.47 & 2.02 & 3.81 & \textbf{3.36} \\\cline{2-9}
        & AdFair-CLIP & \textbf{59.65} & \textbf{86.38} & \textbf{2.27} & \textbf{8.69} & \textbf{0.79} & \textbf{2.93} & 3.54\\    
    \hline
    \multirow{3}{*}{Gender}
        & CLIP  & 57.33 & 86.02 & 6.10 & 6.98 & 2.32 & 3.35 & 2.69 \\
        & FairCLIP  & 59.34 & 86.17 & 3.48 & 6.12 & 2.01 & 3.21 & 2.47 \\\cline{2-9}
        & AdFair-CLIP & \textbf{59.52} & \textbf{86.33} & \textbf{2.18} & \textbf{5.13} & \textbf{1.44} & \textbf{2.61} & \textbf{1.88}\\
    \hline
    \end{tabular}
    }
    \label{tab:chexpert_performance_metrics_linear}
\end{table}
\begin{table}[!h]
    \centering
    \caption{Few-shot (10\%) comparison of AdFair-CLIP with baselines, pre-trained on MIMIC-CXR and fine-tuned on CheXpert Plus, with all scores presented in percentage.}
    \resizebox{\textwidth}{!}{%
    {\fontsize{8pt}{10pt}\selectfont
    \begin{tabular}{|c|l|c|c|c|c|c|c|c|c|}
        \hline
        \textbf{Attribute} & \textbf{Method} & \textbf{Acc} & \textbf{AUC} & \textbf{DPD} & \textbf{DEOdds} & \textbf{GAUC} & \textbf{Inter-AUC} & \textbf{Intra-AUC} \\\hline
        
        \multirow{3}{*}{\centering Race}
        & CLIP  & 55.56 & 84.95 & 6.27 & 11.68 & 2.67 & 3.58 & 4.83 \\
        & FairCLIP & 57.47 & 86.07 & 5.96 & 9.25  & \textbf{1.86} & 3.71 & 3.95 \\\cline{2-9}
        & AdFair-CLIP & \textbf{57.78} & \textbf{86.15} & \textbf{3.73} & \textbf{8.20} & 1.96 & \textbf{2.89} & \textbf{2.74} \\
        \hline
        \multirow{3}{*}{\centering Gender}
        & CLIP  & 55.56 & 84.95 & 6.89 & 10.57 & 2.98 & 3.25 & 4.84 \\
        & FairCLIP & 56.02 & 85.18 & 4.81 & 10.02 & \textbf{1.77} & 3.31 & 3.69 \\\cline{2-9}
        & AdFair-CLIP & \textbf{56.11} & \textbf{85.48} & \textbf{3.31} & \textbf{8.98} & 2.01 & \textbf{2.30} & \textbf{3.29} \\
        \hline
    \end{tabular}
    }
    }
    \label{tab:mimic_performance_metrics_linear}
\end{table}

{\bf AdFair-CLIP vs. CLIP Fairness Benchmarks.} 
Tables~\ref{tab:chexpert_performance_metrics_zeroshot}, \ref{tab:chexpert_performance_metrics_linear}, and \ref{tab:mimic_performance_metrics_linear} demonstrate that AdFair-CLIP consistently achieves the highest fairness and diagnostic performance improvements across most metrics in all evaluated scenarios for race and gender. Notably, DebiasCLIP is only compared in scenario (a) because its frozen strategy makes it equivalent to CLIP in scenarios (b) and (c). To illustrate AdFair-CLIP's superiority in in-domain and out-of-domain generalization, we compute fairness improvements over CLIP by averaging percentage gains across five fairness metrics. In the zero-shot setting, AdFair-CLIP achieves the highest improvements (54.33\% for race and 45.40\% for gender), outperforming DebiasCLIP (36.85\% and 39.94\%) and FairCLIP (33.84\% and 25.39\%). Although fine-tuning and transfer learning introduce additional biases that degrade fairness across all methods, AdFair-CLIP remains robust, achieving 44.14\% (race) and 36.18\% (gender) in the few-shot setting and 31.89\% (race) and 32.16\% (gender) in transfer learning, whereas FairCLIP suffers a marked decline, with race fairness dropping from 33.84\% in the zero-shot setting to 14.14\% in transfer learning and a similar trend observed for gender fairness. These results confirm that AdFair-CLIP's feature-space intervention is more effective and generalizable than similarity-space debiasing and the backbone-frozen fine-tuning strategy.

\subsection{Ablation}

\begin{table}[tbp]
    \centering
    \caption{Vision-only vs. multimodal (vision + text) features for adversarial learning.}
    {\fontsize{8pt}{10pt}\selectfont
   
    \begin{tabular}{|c|l|c|c|c|c|c|c|c|c|}
        \hline
        \multicolumn{9}{|c|}{\textbf{Zeroshot}} \\ \hline
        \textbf{Attribute}& \textbf{Method} & \textbf{Acc} & \textbf{AUC} & \textbf{DPD} & \textbf{DEOdds} & \textbf{GAUC} & \textbf{Inter-AUC} & \textbf{Intra-AUC}\\
        \hline
        \multirow{2}{*}{Race}
        & Vision-only & \textbf{61.06} & 84.83 & 2.13 & 8.16 & 1.36 & 3.15 & 2.98 \\
        & Multimodal & 60.91 & \textbf{85.05} & \textbf{1.87} & \textbf{6.04} & \textbf{0.99} & \textbf{2.46} & \textbf{2.68}\\
        \hline
        \multirow{2}{*}{Gender}
        & Vision-only & 58.44 & 84.17 & 3.09 & 8.80 & \textbf{1.73} & 1.77 & 2.87 \\
        & Multimodal & \textbf{58.77} & \textbf{84.56} & \textbf{2.75} & \textbf{8.69} & 1.82 & \textbf{1.75} & \textbf{2.29}\\
        
        \hline
        \multicolumn{9}{|c|}{\textbf{Linear Evaluation}} \\ \hline
        
        \multirow{2}{*}{Race} 
        & Vision-only & 59.11 & 86.25 & 2.53 & 9.47 & 1.54 & 3.45 & \textbf{3.41}\\
        & Multimodal & \textbf{59.65} & \textbf{86.38} & \textbf{2.27} & \textbf{8.69} & \textbf{0.79} & \textbf{2.93} & 3.54\\    
        \hline
        \multirow{2}{*}{Gender}
        & Vision-only & 59.13 & 86.15 & 3.26 & 5.82 & 1.72 & 3.17 & 2.18 \\
        & Multimodal & \textbf{59.52} & \textbf{86.33} & \textbf{2.18} & \textbf{5.13} & \textbf{1.44} & \textbf{2.61} & \textbf{1.88}\\
        \hline
    \end{tabular}
    }
    \label{tab:comparison_zeroshot}
\end{table}

\noindent{\bf Vision-only vs. Multimodal Features.}  
To examine whether bias is propagated or amplified during the image-text pairing process, we conduct an ablation study comparing the performance of our AdFair-CLIP model on CheXpert Plus when employing vision-only features (\( h_v \)) versus multimodal features (\([h_v; h_u]\)) in the min-max adversarial learning framework. As shown in Table~\ref{tab:comparison_zeroshot}, the multimodal approach achieves superior diagnostic performance and improved fairness metrics compared to the vision-only setting. These results indicate that sensitive information can be implicitly encoded in vision-text alignments, even when textual inputs do not explicitly contain demographic information. Furthermore, joint optimization over multimodal features enables the adversarial framework to more effectively suppress biases introduced through image-text pairing, leading to enhanced fairness and more robust debiased representations.

\section{Conclusion}
Our investigation reveals significant fairness issues in existing SOTA CLIP-based chest X-ray models, highlighting the need for fairness-aware multimodal learning. To address this, we propose AdFair-CLIP, an adversarial framework designed to mitigate demographic biases in CLIP-based chest X-ray diagnosis. By incorporating a minimax adversarial objective in the feature space, our method reduces spurious correlations between sensitive attributes and diagnostic predictions while preserving robust multimodal alignment. Extensive evaluations demonstrate that AdFair-CLIP improves both fairness and diagnostic performance while exhibiting strong generalization in zero-shot and few-shot settings. These findings establish a new benchmark for fairness-aware medical AI in chest X-ray and encourage further research into equitable healthcare applications.
\subsubsection{\discintname}
The authors have no competing interests to declare that are relevant to the content of this article.
%
%
\bibliographystyle{splncs04}
\bibliography{references}
%




\end{document}